\documentclass{sig-alternate-05-2015}

\conferenceinfo{KDD '16,}{August 13 - 17, 2016, San Francisco, CA, USA}
\isbn{978-1-4503-4232-2/16/08}\acmPrice{\$0.00}
\doi{http://dx.doi.org/10.1145/2939672.2939803}

\usepackage{amsmath}
\usepackage{array}
\usepackage{times}
\usepackage{color}
\usepackage{url}
\usepackage{graphicx}
\usepackage{subfigure}
\usepackage{multirow}
\usepackage{epstopdf}
\usepackage{booktabs}

\begin{document}

\widowpenalty=10000
\clubpenalty=10000

\newcolumntype{P}[1]{>{\centering\arraybackslash}p{#1}}

\newcommand{\xhdr}[1]{\paragraph*{\bf {#1}}}

\title{Assessing~Human~Error~Against~a~Benchmark~of~Perfection}

\numberofauthors{3}
\author{
\alignauthor{Ashton Anderson}\\
    \affaddr{Microsoft Research}
    \alignauthor Jon Kleinberg\\
    \affaddr{Cornell University}
    \alignauthor Sendhil Mullainathan\\
    \affaddr{Harvard University}
}

\maketitle

\begin{abstract}
An increasing number of domains are providing us with detailed
trace data on human decisions in settings where we can evaluate
the quality of these decisions via an algorithm.
Motivated by this development, an emerging line of work has
begun to consider whether we can characterize and predict the kinds
of decisions where people are likely to make errors.

To investigate what a general framework for human error prediction
might look like, we focus on a model system with a rich history
in the behavioral sciences: the decisions made by chess players
as they select moves in a game.
We carry out our analysis at a large scale, employing datasets with
several million recorded games, and using {\em chess tablebases}
to acquire a form of ground truth for a
subset of chess positions that have been completely solved by computers
but remain challenging even for the best players in the world.

We organize our analysis around three categories of features that
we argue are present in most settings where the analysis of human error
is applicable: the skill of the decision-maker, the time available
to make the decision, and the inherent difficulty of the decision.
We identify rich structure in all three of these categories of features,
and find strong evidence that in our domain,
features describing the inherent difficulty of an instance are
significantly more powerful than features based on skill or time.

\end{abstract}

\section{Introduction}
\label{sec:intro}

Several rich strands of work in the behavioral sciences
have been concerned with characterizing the nature and sources of human error.
These include the broad of notion of {\em bounded rationality}
\cite{simon-models-of-man} and the subsequent research beginning
with Kahneman and Tversky on heuristics and biases
\cite{tversky-heuristics-biases}.
With the growing availability of large datasets containing millions of
human decisions on fixed, well-defined, real-world tasks,
there is an interesting opportunity to add a new style of inquiry
to this research --- given a large stream of decisions, with
rich information about the context of each decision, can we
algorithmically characterize and predict the instances on which
people are likely to make errors?

This genre of question --- analyzing human errors from
large traces of decisions on a fixed task --- also has an interesting
relation to the canonical set-up in machine learning applications.
Typically, using instances of decision problems together
with ``ground truth'' showing the correct decision, an algorithm
is trained to produce the correct decisions in a large fraction of instances.
The analysis of human error, on the other hand, represents a
twist on this formulation: given instances of a task in which we have
both the correct decision {\em and} a human's decision,
the algorithm is trained to recognize future instances on which
the human is likely to make a mistake.
Predicting human error from this type of trace data has a history in
human factors research \cite{kirwan-human-reliability,salvendy-human-error},
and a nascent line of work has begun to apply current
machine-learning methods to the question
\cite{lakkaraju-bail-working-paper,lakkaraju-sdm15-eval}.

\xhdr{Model systems for studying human error}
As the investigation of human error using large datasets grows
increasingly feasible, it becomes useful to understand which styles
of analysis will be most effective.
For this purpose, as in other settings, there is enormous value in
focusing on model systems where one has exactly the data necessary
to ask the basic questions in their most natural formulations.

What might we want from such a model system?
\begin{itemize}
\item[(i)] It should consist of
a task for which the context of the human decisions has
been measured as thoroughly as possible, and in a very large number of
instances, to provide the training data for an algorithm to analyze errors.
\item[(ii)] So that the task is non-trivial, it should be challenging even for
highly skilled human decision-makers.
\item[(iii)] Notwithstanding the previous point (ii), the ``ground truth'' ---
the correctness of each candidate decision --- should be feasibly computable
by an algorithm.
\end{itemize}

Guided by these desiderata, we focus in this paper on chess
as a model system for our analysis.
In doing so, we are proceeding by analogy with a long line of
work in behavioral science
using chess as a model for
human decision-making
\cite{charness-psych-chess-review,chase-simon-perception-chess,degroot-thought-choice-chess}.
Chess is a natural domain for such investigations,
since it presents a human player with a sequence of concrete
decisions --- which move to play next --- with the property that
some choices are better than others.
Indeed, because chess provides data on
hard decision problems in such a pure fashion,
it has been described as the
``drosophila of psychology''
\cite{didierjean-chess-since-degroot,halpern-wai-scrabble}.
(It is worth noting our focus here on {\em human} decisions
in chess, rather than on designing algorithms to play chess
\cite{newborn-computer-chess}.  This latter problem has also,
of course, generated a rich literature, along with a closely
related tag-line as the
``drosophila of artificial intelligence''
\cite{mccarthy-chess-drosophila}.)

\xhdr{Chess as a model system for human error}
Despite the clean formulation of the decisions made by
human chess players, we still must resolve a set of conceptual
challenges if our goal
is to assemble a large corpus of chess moves with ground-truth
labels that classify certain moves as errors.
Let us consider three initial ideas for how we might go about this,
each of which is lacking in some crucial respect for our purposes.

First, for most of the history of human decision-making
research on chess, the emphasis
has been on focused laboratory studies at small scales in which the correct
decision could be controlled by design
\cite{charness-psych-chess-review}.
In our list of desiderata, this means that point (iii), the availability
of ground truth, is well under control, but a significant aspect of
point (i) --- the availability of a vast number of instances --- is
problematic due to the necessarily small scales of the studies.

A second alternative would be to make use of two important computational
developments in chess --- the availability of databases with millions
of recorded chess games by strong players; and the fact that
the strongest chess programs ---
generally referred to as {\em chess engines} ---
now greatly outperform even the best human players in the world.
This makes it possible to analyze the moves of strong human players,
in a large-scale fashion, comparing their choices to those of an engine.
This has been pursued very effectively in the last several years by
Biswas and Regan
\cite{biswas-regan-icmla15,biswas-regan-icaart15,regan-biswas-cig13};
they have used the approach to derive interesting insights including
proposals for how to estimate the depth at which human players are
analyzing a position.

For the current purpose of assembling a corpus with ground-truth
error labels, however, engines present a set of challenges.
The basic difficulty is that
even current chess engines are far from being able to
provide guarantees regarding the best move(s) in a given position.
In particular, an engine may prefer move $m$ to $m'$
in a given position, supplementing this preference with a heuristic numerical
evaluation, but $m'$ may ultimately lead to the same result in the game,
both under best play and under typical play.
In these cases, it is hard to say that choosing $m'$ should be
labeled an error.
More broadly, it is difficult to find a clear-cut rule mapping
an engine's evaluations to a determination of human error, and
efforts to label errors this way would represent a complex mixture of the
human player's mistakes and the nature of the engine's evaluations.

Finally, a third possibility is to go back to the definition of chess as a
deterministic game with two players
(White and Black) who engage in alternating moves, and with
a game outcome that is either (a) a win for White, (b) a win for Black,
or (c) a draw.
This means that from any position, there is a well-defined notion
of the outcome with respect to optimal play by both sides --- in
game-theoretic terms, this is the {\em minimax value} of the position.
In each position, it is the case that White wins with best play,
or Black wins with best play, or it is a draw with best play,
and these are the three possible minimax values for the position.

This perspective provide us with a clean option for formulating
the notion of an error, namely the direct game-theoretic definition:
a player has committed an error if their move worsens the minimax
value from their perspective.
That is, the player had a forced win before making their move but
now they don't; or the player had a forced draw before making their move
but now they don't.
But there's an obvious difficulty with this route, and it's a
computational one: for most chess positions,
determining the minimax value is hopelessly beyond the power of
both human players and chess engines alike.

We now discuss the approach we take here.

\xhdr{Assessing errors using tablebases}
In our work, we use minimax values by leveraging a further
development in computer chess ---
the fact that chess has been solved
for all positions with at most $k$ pieces on the board,
for small values of $k$
\cite{url-lomonosov-tablebases,url-nalimov-tablebases,kopec-advances-man-machine}.
(We will refer to such positions as {\em $\leq$$k$-piece positions}.)
Solving these positions
has been accomplished not by forward construction of the chess game tree,
but instead by simply
working backward from terminal positions with a
concrete outcome present on the board
and filling in all other minimax values by dynamic programming until all possible $\leq$$k$-piece positions have been enumerated.
The resulting solution for all $\leq$$k$-piece positions
is compiled into an object called a {\em $k$-piece tablebase}, which lists
the game outcome with best play for each of these positions.
The construction of tablebases has been a topic of interest since
the early days of computer chess \cite{kopec-advances-man-machine},
but only with recent developments
in computing and storage have truly large tablebases been feasible.
Proprietary tablebases with $k = 7$ have been built, requiring
in excess of a hundred terabytes of storage \cite{url-lomonosov-tablebases};
tablebases for $k = 6$ are much more manageable, though still very large
\cite{url-nalimov-tablebases}, and we focus on the case of $k = 6$ in what
follows.\footnote{There are some intricacies in how tablebases interact
with certain rules for draws in chess, particularly
{\em threefold-repetition} and {\em the 50-move rule}, but since these
have essentially negligible impact on our use of tablebases in the present
work, we do not go into further details here.}

Tablebases and traditional chess engines are thus very different objects.
Chess engines produce strong moves for arbitrary positions,
but with no absolute guarantees on move quality in most cases;
tablebases, on the other hand,
play perfectly with respect to the game tree --- indeed,
effortlessly, via table lookup --- for
the subset of chess containing at most $k$ pieces on the board.

Thus, for arbitrary $\leq$$k$-piece positions, we can determine
minimax values, and so we can obtain a large corpus of chess moves
with ground-truth error labels:
Starting with a large database of recorded chess games,
we first restrict to the subset of $\leq$$k$-piece positions,
and then we label a move as an error if and only it worsens
the minimax value from the perspective of the player making the move.
Adapting chess terminology
to the current setting, we will refer to such an instance
as a {\em blunder}.

This is our model system for analyzing human error; let us now check
how it lines up with desiderata (i)-(iii) for a model system
listed above.
Chess positions with at most $k = 6$ pieces arise relatively frequently
in real games, so we are left with many instances even after
filtering a database of games to restrict
to only these positions (point (i)).
Crucially, despite their simple structure, they can induce
high error rates by amateurs and non-trivial
error rates even by the best players in the world;
in recognition of the inherent challenge they contain,
textbook-level treatments of chess devote a significant fraction
of their attention to these positions \cite{fine-basic-chess-endings}
(point (ii)).
And they can be evaluated perfectly by tablebases (point (iii)).

Focusing on $\leq$$k$-piece positions has an additional benefit, made possible by a combination of tablebases and the recent availability of databases with millions of recorded chess games.
The most frequently-occurring of these positions arise in our
data thousands of times.  As we will see, this means that for some
of our analyses, we can control for the exact position on the board
and still have enough instances to observe meaningful variation.
Controlling for the exact position is not generally
feasible with arbitrary positions arising in the middle of a chess game,
but it becomes possible with the scale of data we now have,
and we will see that in this case it yields interesting and in some
cases surprising insights.

Finally, we note that our definition of blunders, while concrete and
precisely aligned with the minimax value of the game tree, is not the only
definition that could be considered even using tablebase evaluations.
In particular,
it would also be possible to consider ``softer'' notions of blunders.
Suppose for example that a player is choosing between moves $m$ and $m'$,
each leading to a position whose minimax value is a draw, but suppose
that the position arising after $m$ is more difficult for the opponent,
and produces a much higher empirical probability that the opponent
will make a mistake at some future point and lose.
Then it can be viewed as a
kind of blunder, given these empirical probabilities,
to play $m'$ rather than the more challenging $m$.
This is sometimes termed {\em speculative play}
\cite{jansen-speculative-play}, and it can be thought of primarily as a
refinement of the coarser minimax value.
This is an interesting extension, but for our
work here we focus on the purer notion of blunders based on the minimax value.

\section{Setting Up the Analysis}
\label{sec:features}

In formulating our analysis, we begin from the premise that
for analyzing error in human decisions,
three crucial types of features are the following:
\begin{itemize}
\itemsep0em
\item[(a)] the skill of the decision-maker;
\item[(b)] the time available to make the decision; and
\item[(c)] the inherent difficulty of the decision.
\end{itemize}
Any instance of the problem will implicitly or explicitly contain
features of all three types: an individual of a particular level
of skill is confronting a decision of a particular difficulty,
with a given amount of time available to make the decision.

In our current domain, as in any other setting where
the question of human error is relevant, there are a number
of basic genres of question that we would like to ask.
These include the following.

\begin{itemize}
\itemsep0em
\item For predicting whether an error will be committed in a given
instance, which types of features (skill, time, or difficulty)
yield the most predictive power?
\item In which kinds of instances does greater skill confer
the largest relative benefit?  Is it for more difficult decisions (where
skill is perhaps most essential) or for easier ones (where
there is the greatest room to realize the benefit)?
Are there particular kinds of instances where skill does not
in fact confer an appreciable benefit?
\item An analogous set of questions for time in place of skill:
In which kinds of instances does greater time for the decision
confer the largest benefit?  Is additional time more beneficial for hard decisions or easy ones?  And are there
instances where additional time does not reduce the error rate?
\item Finally, there are natural questions about the interaction
of skill and time: is it higher-skill or lower-skill decision-makers
who benefit more from additional time?
\end{itemize}

These questions motivate our analyses in the
subsequent sections.  We begin by discussing how features of
all three types (skill, time, and difficulty)
are well-represented in our domain.

Our data comes from two large databases of recorded chess games.
The first is a corpus of approximately 200 million games from
the Free Internet Chess Server (FICS), where amateurs play each other on-line\footnote{This data is publicly available at {\url{ficsgames.org}}.}.
The second is a corpus of approximately 1 million games played
in international tournaments by the strongest players in the world.
We will refer to the first of these as the FICS dataset, and the
second as the GM dataset.  (GM for ``grandmaster,'' the highest
title a chess player can hold.)
For each corpus, we extract all occurrences of
$\leq$$6$-piece positions from all of the games;
we record the move made in the game from each occurrence of each position,
and use a tablebase to evaluate all possible moves from the position
(including the move that was made).
This forms a single instance for our analysis.
Since we are interested in studying errors, we exclude all instances in which the player to move is in a theoretically losing position --- where the opponent has a direct path to checkmate --- because there are no blunders in losing positions (the minimax value of the position is already as bad as possible for the player to move).
There are 24.6 million (non-losing) instances in the FICS dataset, and 880,000 in the GM dataset.

We now consider how feature types (a), (b), and (c) are
associated with each instance.
First, for skill, each chess player in the data has a numerical rating,
termed the {\em Elo rating},
based on their performance in the games they've played
\cite{elo-chess-ratings,herbrich-trueskill}.
Higher numbers indicate stronger players, and to get a rough
sense of the range: most amateurs have ratings in the range 1000-2000,
with extremely strong amateurs getting up to 2200-2400;
players above 2500-2600 belong to a rarefied group of the world's best;
and at any time there are generally about fewer than five
people in the world above 2800.
If we think of a game outcome in terms of points, with 1 point for
a win and 0.5 points for a draw, then the Elo rating system has the property
that when a player is paired with someone 400 Elo points lower, their
expected game outcome is approximately $0.91$
points --- an enormous advantage.\footnote{In general,
the system is designed so that when the rating
difference is $400d$, the expected score for the higher-ranked player under the Elo system is $1 / (1 + 10^{-d})$.}

For our purposes, an important feature of Elo ratings is the fact
that a single number has empirically proven so powerful at predicting
performance in chess games.
While ratings clearly cannot contain all the information about
players' strengths and weaknesses, their effectiveness in practice
argues that we can reasonably use a player's rating as a single
numerical feature that approximately represents their skill.

With respect to temporal information,
chess games are generally played under time limits of the form,
``play $x$ moves in $y$ minutes'' or
``play the whole game in $y$ minutes.''
Players can choose how they use this time, so on each move they
face a genuine decision about how much of their remaining allotted
time to spend.
The FICS dataset contains the amount of time remaining in the game
when each move was played (and hence the amount of time spent on
each move as well); most of the games in the FICS dataset are played
under extremely rapid time limits, with a large fraction of them
requiring that the whole game be played in 3 minutes for each player.
To avoid variation arising from the game duration, we focus on this
large subset of the FICS data consisting exclusively of games
with 3 minutes allocated to each side.

Our final set of features will be designed to quantify
the difficulty of the position on the board --- i.e. the extent to
which it is hard to avoid selecting a move that constitutes a blunder.
There are many ways in which one could do this, and we are guided
in part by the goal of developing features that are less domain-specific
and more applicable to decision tasks in general.
We begin with perhaps the two most basic parameters, analogues
of which would be present in any setting with discrete choices and a discrete
notion of error --- these are the number of legal moves in the position,
and the number of these moves that constitute blunders.
Later, we will also consider a general family of parameters that
involve looking more deeply into the search tree, at moves beyond
the immediate move the player is facing.

To summarize,
in a single instance in our data, a player of a given rating,
with a given amount of time remaining in the game, faces a specific position on
the board, and we ask whether the move they select is a
blunder.
We now explore how our different types of features provide information
about this question, before turning to the general problem of prediction.

\section{Fundamental Dimensions}
\label{sec:basic}

\subsection{Difficulty}
\label{subsec:difficulty}

We begin by considering a set of basic features that help quantify
the difficulty inherent in a position.
There are many features we could imagine employing that are
highly domain-specific to chess, but our primary interest is in whether
a set of relatively generic features can provide non-trivial predictive value.

Above we noted that in any setting with discrete choices,
one can always consider the total number of available choices,
and partition these into the number that constitute blunders and
the number that do not constitute blunders.
In particular, let's say that in a given chess position $P$,
there are $n(P)$ legal moves available --- these are the possible choices ---
and of these, $b(P)$ are blunders, in that they lead to a position
with a strictly worse minimax value.
Note that it is possible to have $b(P) = 0$, but we exclude these positions because it is impossible to blunder. Also, by the definition
of the minimax value, we must have $b(P) \leq n(P) - 1$; that is,
there is always at least one move that preserves the minimax value.

\begin{figure}[t]
\centering
\includegraphics[width=0.4\textwidth]{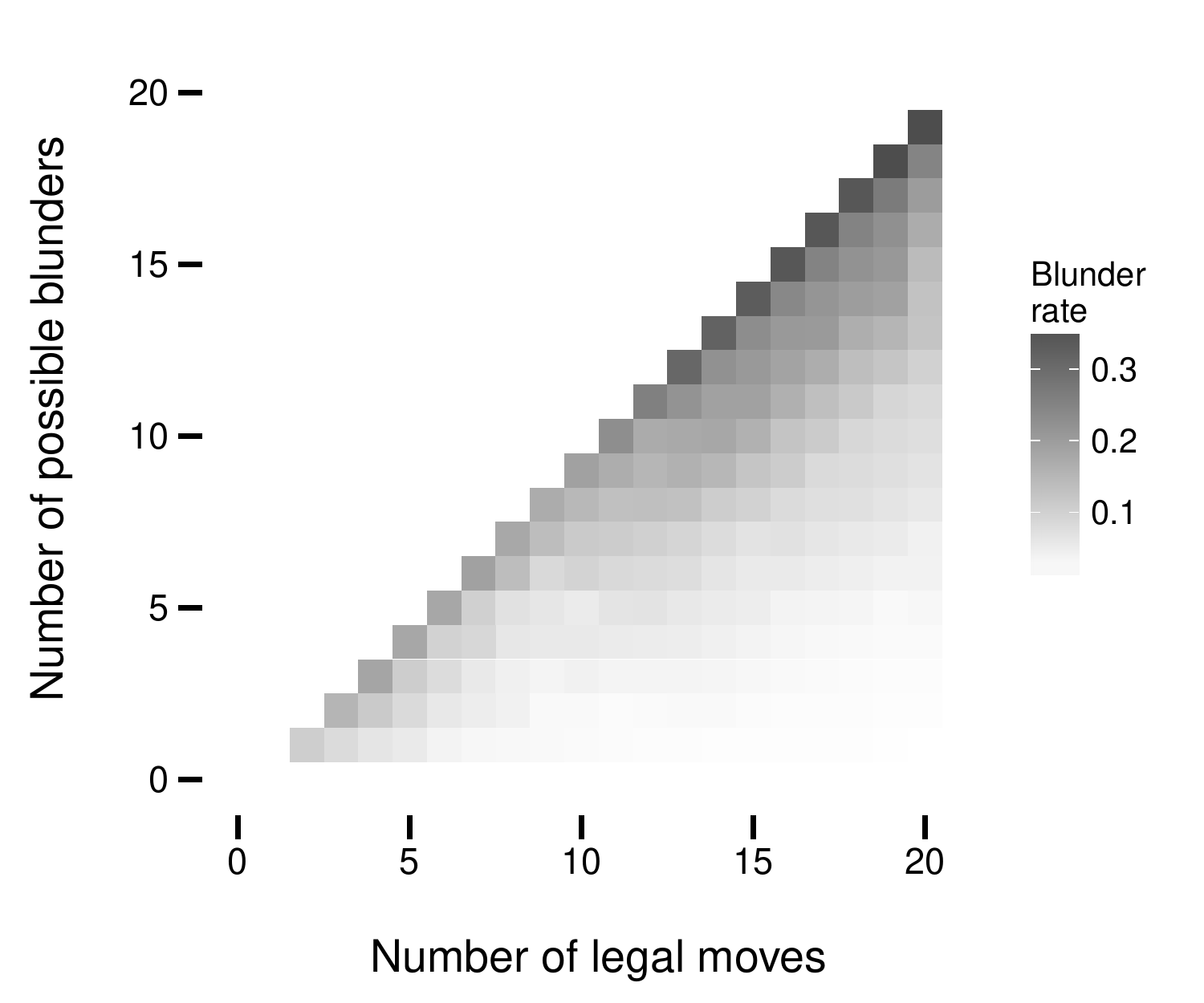}
\vspace{-2mm}
\caption{A heat map showing the empirical blunder rate as a
function of the two variables $(n(P),b(P))$, for the FICS dataset.}
\label{fig:blunder-rate-heat-map}
\vspace{-3mm}
\end{figure}

A global check of the data reveals an interesting bimodality
in both the FICS and GM datasets:
positions with $b(P) = 1$ and positions with $b(P) = n(P) - 1$
are both heavily represented.  The former correspond to positions in which
there is a unique blunder, and the latter correspond to
positions in which there is a unique correct move to preserve the
minimax value.
Our results will cover the full range of $(n(P),b(P))$ values,
but it is useful to know that both of these extremes are
well-represented.

Now, let us ask what the empirical blunder rate looks like as a
bivariate function of this pair of variables $(n(P),b(P))$. Over all instances in which the underlying position $P$ satisfies
$n(P) = n$ and $b(P) = b$, we define $r(n,b)$ to be the fraction
of those instances in which the player blunders.
How does the empirical blunder rate vary in $n(P)$ and $b(P)$?
It seems natural to suppose that for fixed $n(P)$, it should generally
increase in $b(P)$, since there are more possible blunders to make.
On the other hand, instances with $b(P) = n(P) - 1$ often correspond to chess positions in which the only non-blunder is ``obvious'' (for example, if there is only one way to recapture a piece), and so one might conjecture that the empirical blunder rate will be lower for this case.

In fact, the empirical blunder rate is generally monotone in $b(P)$, as shown by the heatmap representation of $r(n,b)$ in
Figure \ref{fig:blunder-rate-heat-map}.
(We show the function for the FICS data; the function for the
GM data is similar.)
Moreover, if we look at the heavily-populated line
$b(P) = n(P) - 1$, the blunder rate is increasing
in $n(P)$; as there are more blunders to compete with the unique
non-blunder, it becomes correspondingly harder to make the right choice.

\begin{figure}[t]
\centering
\includegraphics[width=0.23\textwidth]{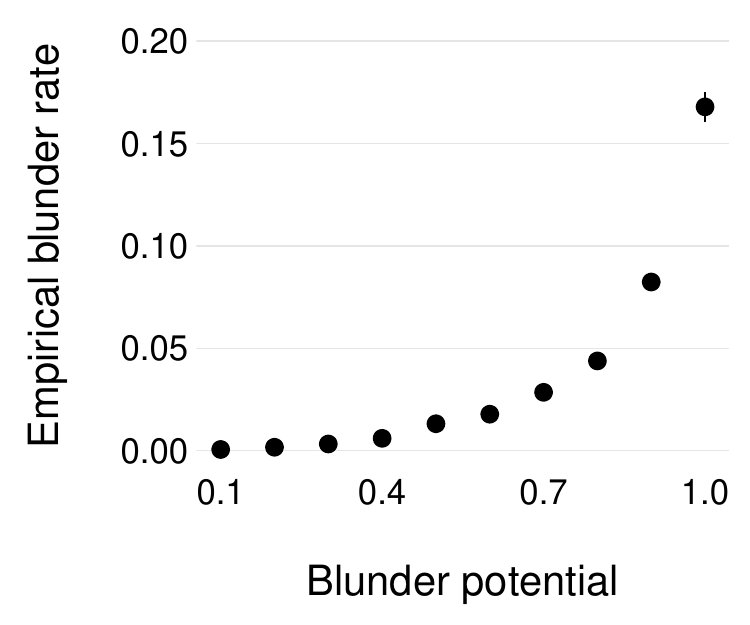}
\includegraphics[width=0.23\textwidth]{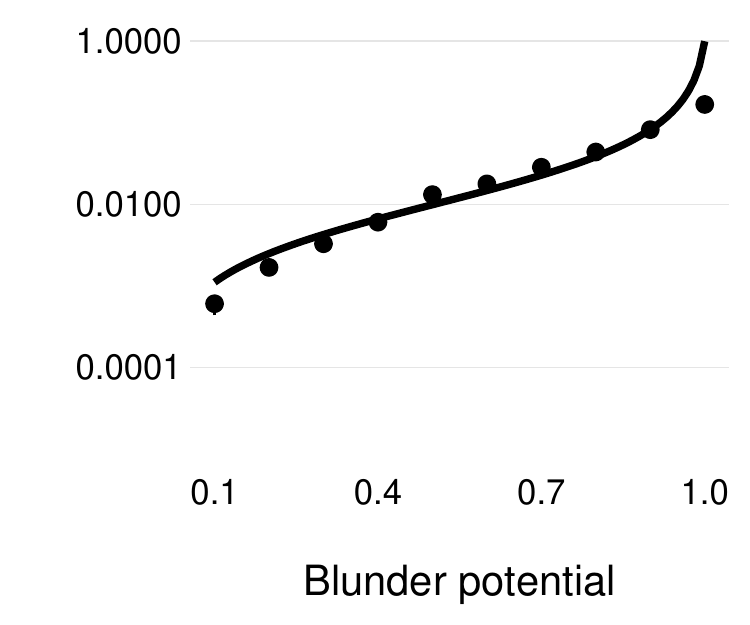}\\
(a) GM data \\
\includegraphics[width=0.23\textwidth]{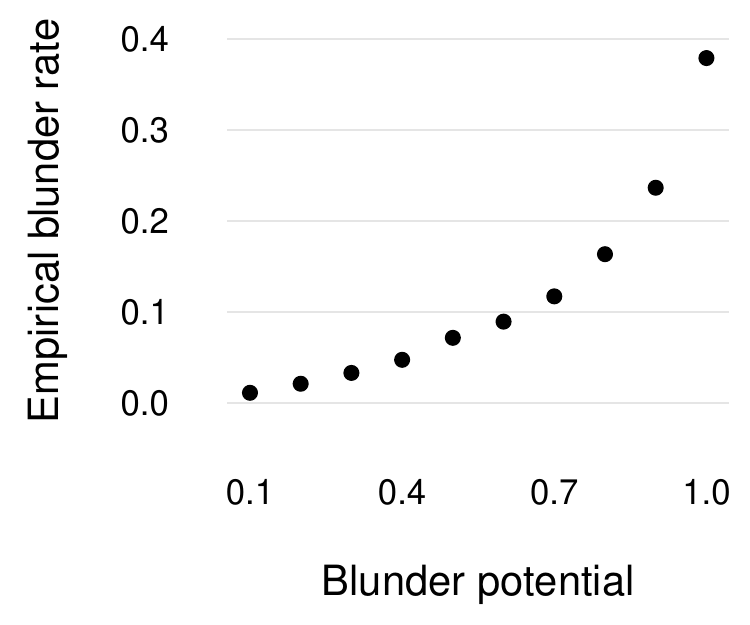}
\includegraphics[width=0.23\textwidth]{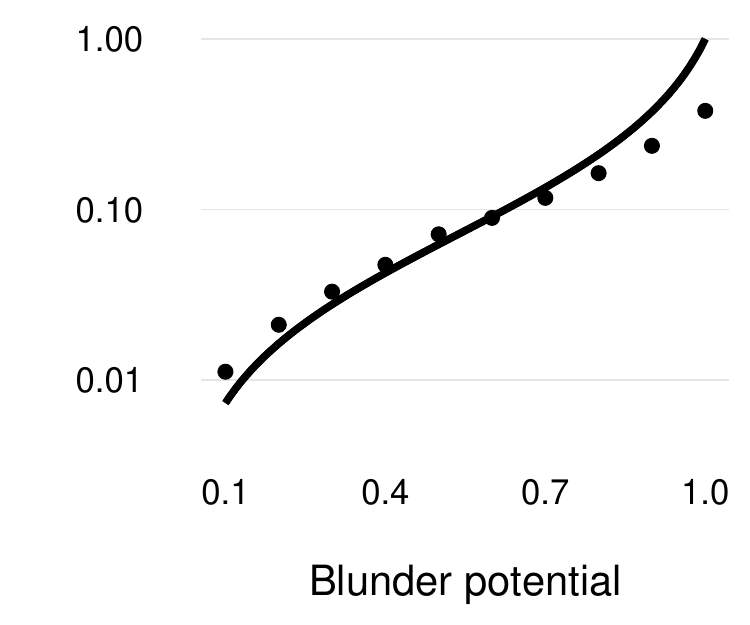} \\
(b) FICS data
\vspace{-2mm}
\caption{The empirical blunder rate as a function of the blunder potential,
shown for both the GM and the FICS data.  On the left are standard axes,
on the right are logarithmic y-axes.  The plots on the right also show
an approximate fit to the $\gamma$-value defined in
Section \ref{subsec:difficulty}.}
\label{fig:blunder-rate-blunder-potential}
\vspace{-3mm}
\end{figure}

\xhdr{Blunder Potential}
Given the monotonicity we observe, there is an informative way to
combine $n(P)$ and $b(P)$: by simply taking their 
ratio $b(P) / n(P)$.  This quantity, which we term the
{\em blunder potential} of a position $P$ and denote
$\beta(P)$, is the answer to the question,
``If the player selects a move uniformly at random,
what is the probability that they will blunder?''.
This definition will prove useful in many of the analyses to follow.
Intuitively, we can think of it as a direct measure of the
{\em danger} inherent in a position, since it captures the
relative abundance of ways to go wrong.

In Figure \ref{fig:blunder-rate-blunder-potential} we plot the function $y = r(x)$, the proportion of blunders in instances with $\beta(P) = x$, for both our GM and FICS datasets on linear as well as logarithmic $y$-axes. The striking regularity of the $r(x)$ curves shows how strongly the availability of potential mistakes translates into actual errors. One natural starting point for interpreting this relationship is to note that
if players were truly selecting their moves uniformly at random,
then these curves would lie along the line $y = x$.
The fact that they lie below this line indicates that
in aggregate players are preferentially selecting non-blunders,
as one would expect.
And the fact that the curve for the GM data lies much further below
$y = x$ is a reflection of the much greater skill of the players
in this dataset, a point that we will return to shortly.

\xhdr{The $\gamma$-value}
We find that a surprisingly simple model qualitatively captures the shapes
of the curves in Figure \ref{fig:blunder-rate-blunder-potential} quite well.
Suppose that instead of selecting a move uniformly at random,
a player selected from a biased distribution in which they were
preferentially $c$ times more likely to select a non-blunder than a blunder,
for a parameter $c > 1$.

\begin{figure}[t]
\centering
\includegraphics[width=0.3\textwidth]{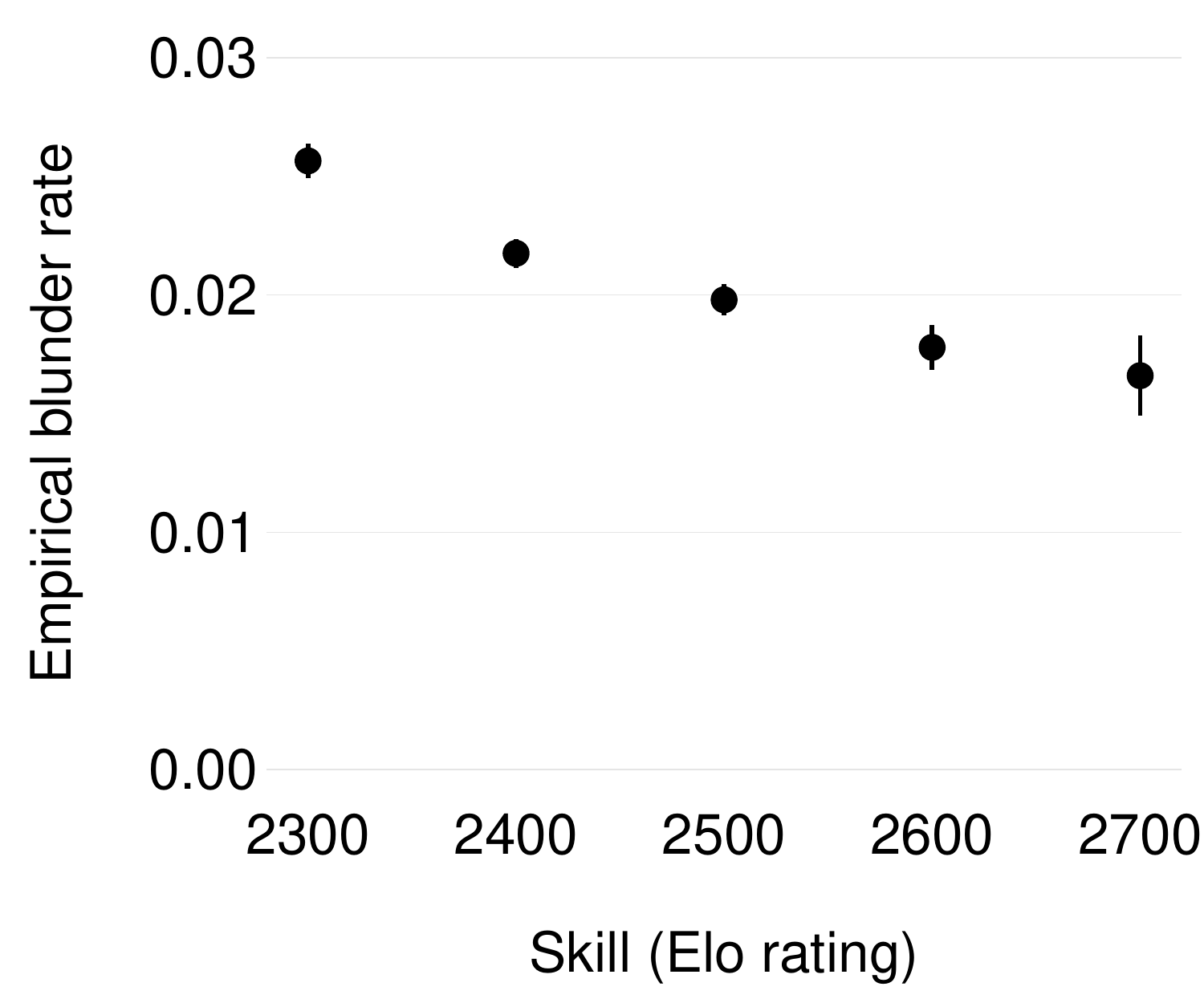} 
\includegraphics[width=0.3\textwidth]{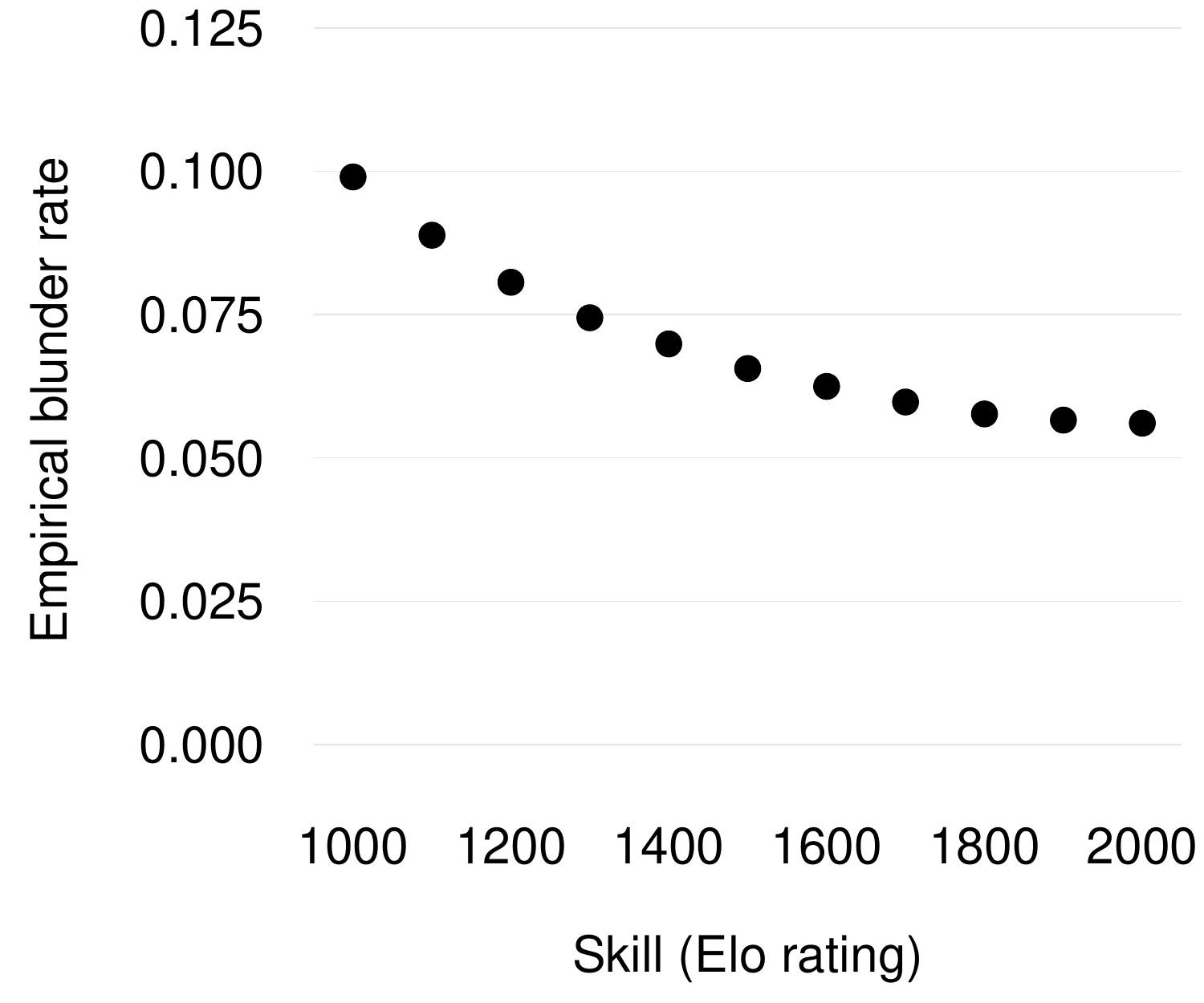} 
\vspace{-2mm}
\caption{The empirical blunder rate as a function of player rating,
shown for both the (top) GM and (bottom) FICS data.}
\label{fig:blunder-rate-rating}
\end{figure}

If this were the true process for move selection, then the
empirical blunder rate of a position $P$ would be
$$\gamma_c(P) = \frac{b(P)}{c (n(P) - b(P)) + b(P)}.$$
We will refer to this as the {\em $\gamma$-value} of the position $P$,
with parameter $c$.
Using the definition of the blunder potential $\beta(P)$
to write $b(P) = \beta(P) n(P)$,
we can express the $\gamma$-value directly as a function of the
blunder potential:
$$\gamma_c(P)
= \frac{\beta(P) n(P)}{c (n(P) - \beta(P) n(P)) + \beta(P) n(P)}
= \frac{\beta(P)}{c - (c-1) \beta(P)}.$$
We can now find the value of $c$ for which $\gamma_c(P)$ best approximates the empirical curves
in Figure \ref{fig:blunder-rate-blunder-potential}.
The best-fit values of $c$ are
$c \approx 15$ for the FICS data and $c \approx 100$ for the GM data,
again reflecting the skill difference between the two domains.
These curves are shown superimposed on
the empirical plot in the figure (on the right, with
logarithmic $y$-axes).

We note that in game-theoretic terms the $\gamma$-value can
be viewed as a kind of {\em quantal response}
\cite{mckelvey-quantal-resp-extensive},
in which players in a game select among alternatives with a probability
that decreases according to a particular function of the alternative's payoff.
Since the minimax value of the position corresponds to the game-theoretic
payoff of the game in our case, a selection rule that probabilistically
favors non-blunders over blunders can be viewed as following this principle.
(We note that our functional form cannot be directly mapped onto
standard quantal response formulations.  The standard formulations
are strictly monotonically decreasing in payoff, whereas we have
cases where two different blunders can move the minimax value by
different amounts --- in particular, when a win changes to a draw
versus a win changes to a loss --- and we treat these the same
in our simple formulation of the $\gamma$-value.)

\subsection{Skill}

\begin{figure}[t]
\centering
\includegraphics[width=0.42\textwidth]{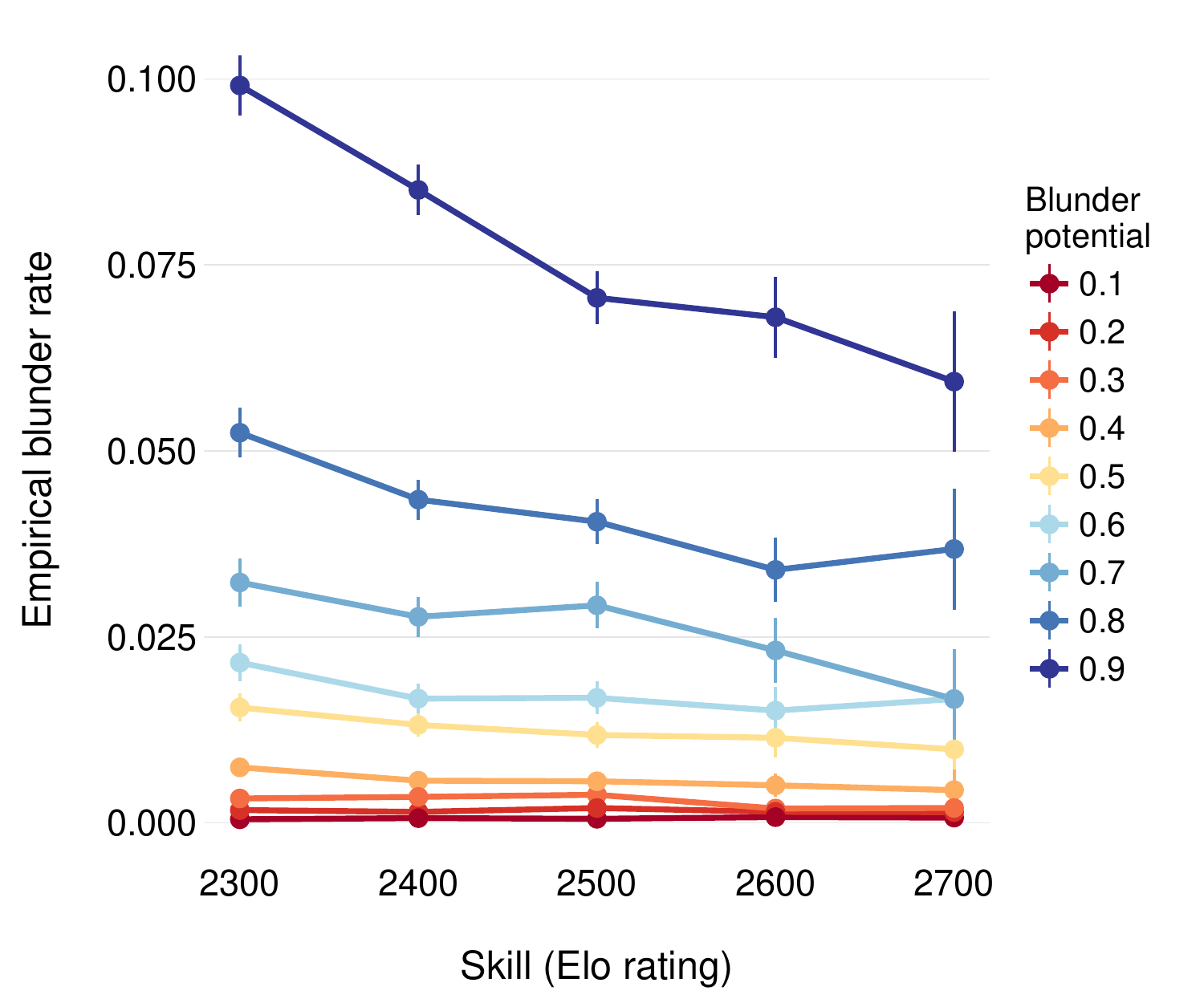}
\vspace{-2mm}

\centering
\includegraphics[width=0.42\textwidth]{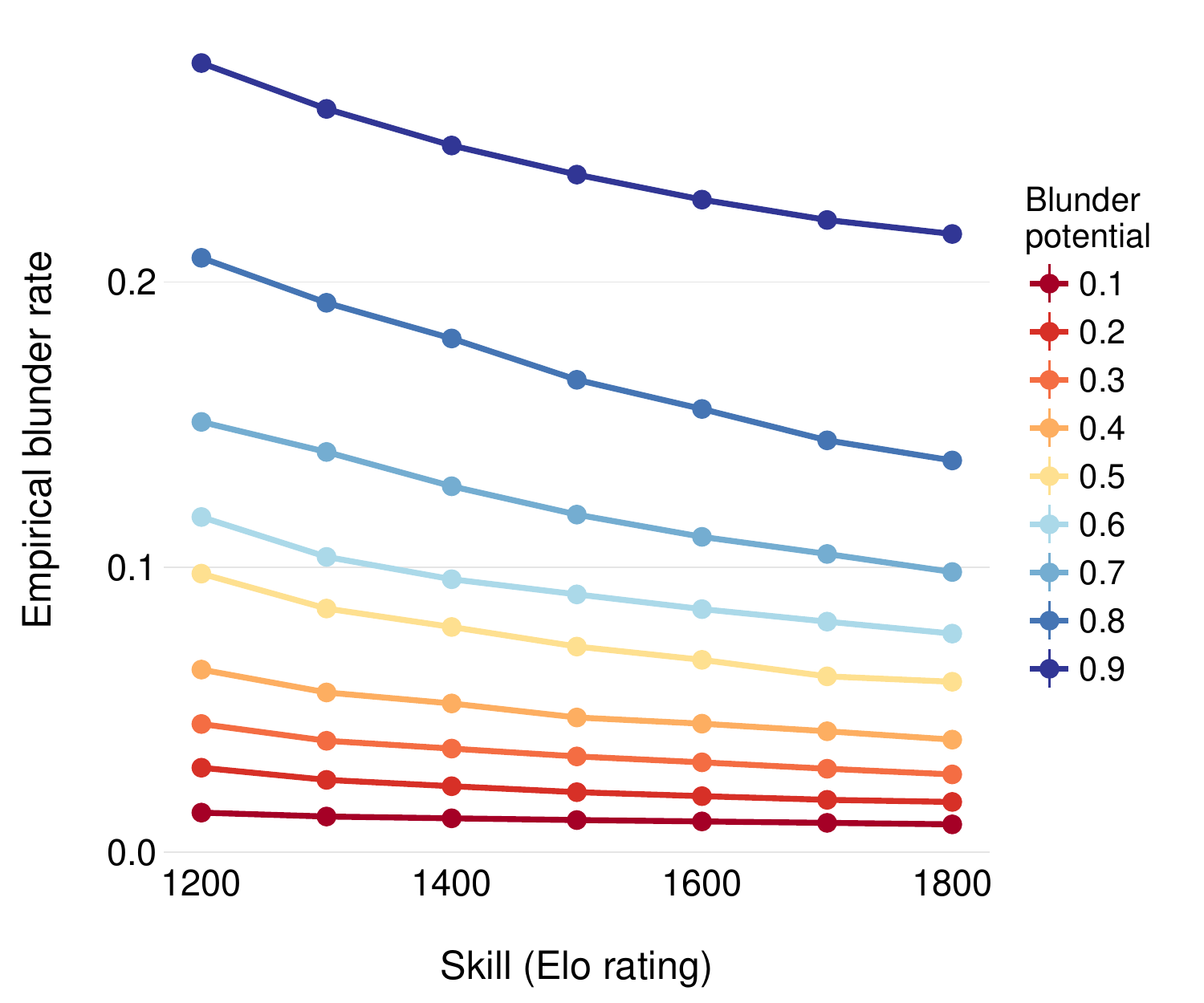}
\vspace{-2mm}
\caption{The empirical blunder rate as a function of Elo rating in the (top) GM and (bottom)
FICS data, for positions with fixed values of the blunder potential.}
\label{fig:blunder-rate-potential-rating}
\vspace{-3mm}
\end{figure}

A key focus in the previous subsection was to understand
how the empirical blunder rate varies as a function of parameters
of the instance.  Here we continue this line of inquiry, with
respect to the skill of the player in addition to the difficulty
of the position.

Recall that a player's {\em Elo rating}
is a function of the outcomes of the games they've played,
and is effective in practice for predicting the outcomes of a
game between two rated players \cite{elo-chess-ratings}.
It is for this reason that we use a player's rating as a proxy
for their skill.
However, given that ratings are determined by which games a player
wins, draws, or loses, rather than by the extent to which they blunder
in $\leq$$6$-piece positions, a first question is whether the
empirical blunder rate in our data shows a clean dependence
on rating.

In fact it does.
Figure \ref{fig:blunder-rate-rating} shows the empirical blunder rate
$f(x)$ averaged over all instances in which the player has rating $x$.
The blunder rate declines smoothly with rating for both the GM and FICS
data, with a flattening of the curve at higher ratings.

\xhdr{The Skill Gradient}
We can think of the downward slope in Figure \ref{fig:blunder-rate-rating} as
a kind of {\em skill gradient}, showing the reduction in blunder rate
as skill increases.
The steeper this reduction is in a given setting,
the higher the empirical benefit of skill in reducing error.

It is therefore natural to ask how the skill gradient varies across
different conditions in our data.
As a first way to address this, we take each possible value
of the blunder potential $\beta$ (rounded
to the nearest multiple of $0.1$), and define the function
$f_\beta(x)$ to be the empirical error rate of players of rating $x$
in positions of blunder potential $\beta$.
Figure \ref{fig:blunder-rate-potential-rating} shows plots of these
curves for $\beta$ equal to each multiple of $0.1$,
for both the GM and FICS datasets.

\begin{figure}[t]
\centering
\includegraphics[width=0.48\textwidth]{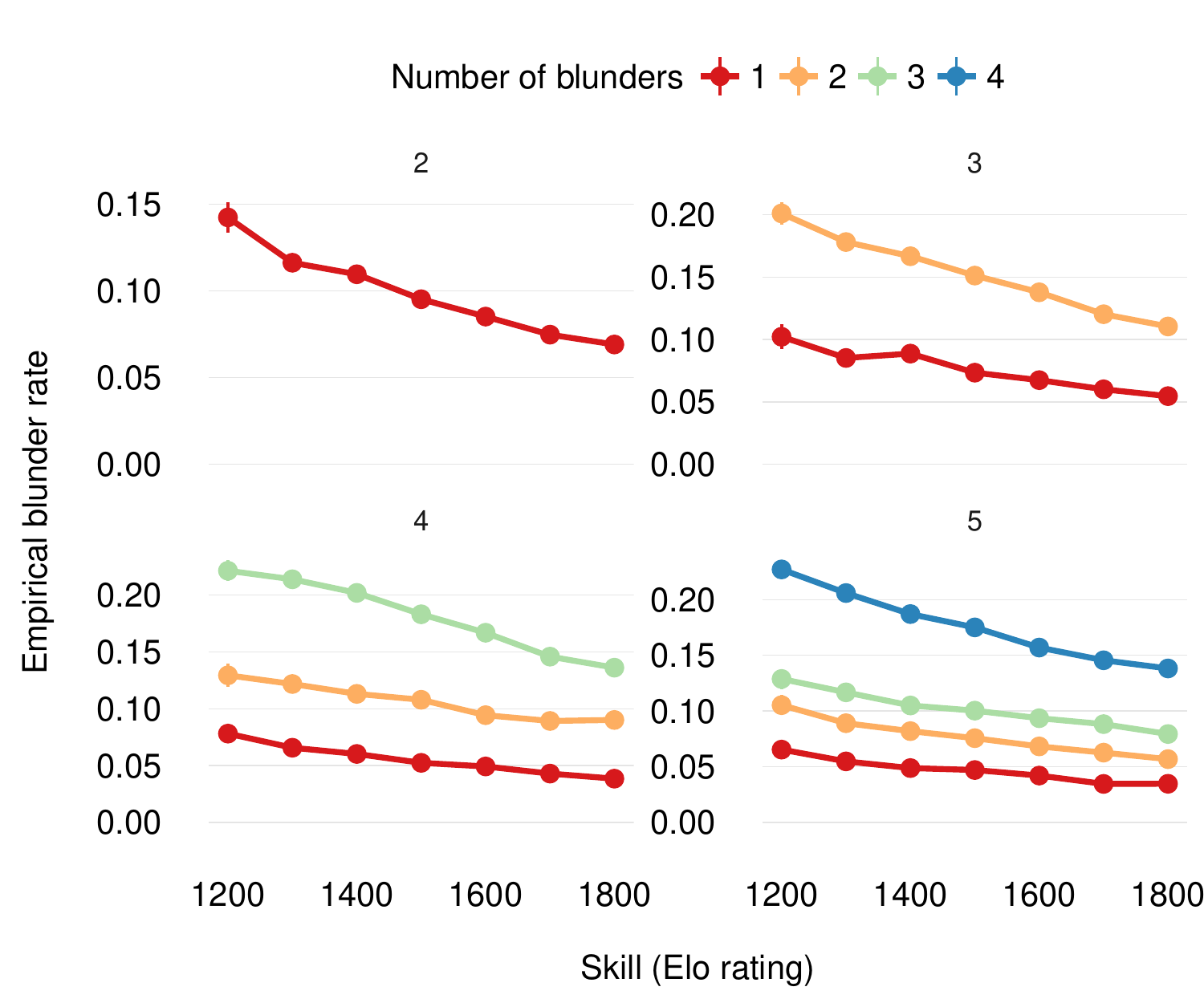}
\caption{The empirical blunder rate as a function of Elo rating in
the FICS data, for positions with fixed values of $(n(P),b(P))$.}
\label{fig:skill-gradient}
\vspace{-3mm}
\end{figure}

\begin{figure}[t]
\centering
\hspace*{-0.03\textwidth}
\includegraphics[width=0.50\textwidth]{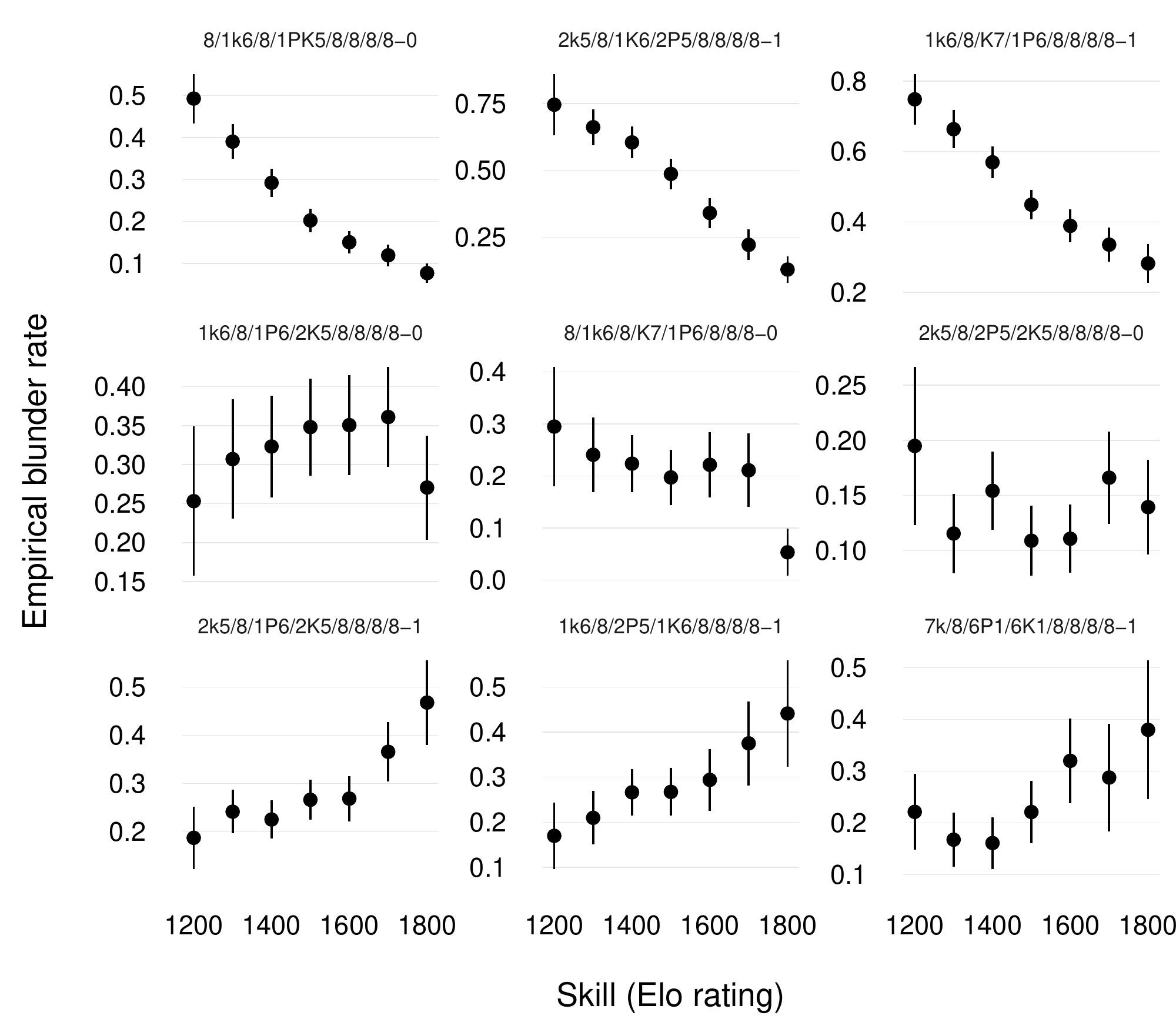} \\
\caption{The empirical blunder rate as a function of Elo rating,
for a set of frequently-occurring positions.}
\label{fig:skill-gradient-position}
\end{figure}

We observe two properties of these curves.
First, there is remarkably little variation among the curves. When viewed on a logarithmic y-axis the curves are almost completely parallel, indicating the same rate of proportional decrease across all blunder potentials.

A second, arguably more striking, property is
how little the curves overlap in their ranges of $y$-values.
In effect, the curves form a kind of ``ladder'' based on
blunder potential:
for every value of the discretized blunder potential,
every rating in 1200-1800 range on FICS has a lower empirical blunder rate
at blunder potential $\beta$ than the best of these ratings at
blunder potential $\beta + 0.2$.
In effect, each additional $0.2$ increment in blunder potential
contributes more, averaging over all instances,
to the aggregate empirical blunder rate than
an additional 600 rating points, despite the fact that
600 rating points represent a vast difference in chess performance.
We see a similar effect for the GM data, where small increases
in blunder potential have a greater effect on blunder rate
than the enormous difference between a rating of 2300 and a rating of 2700.
(Indeed, players rated 2700 are making errors at a greater rate in
positions of blunder potential 0.9 than players rated 1200 are making
in positions of blunder potential 0.3.)
And we see the same effects when we separately fix the numerator and
denominator that constitute the blunder potential, $b(P)$ and $n(P)$,
as shown in Figure \ref{fig:skill-gradient}.

To the extent that this finding runs counter to our intuition,
it bears an interesting relation to the
{\em fundamental attribution error} --- the tendency to attribute
differences in people's performance to differences in their
individual attributes, rather than to differences in the situations
they face \cite{jones-fundamental-attrib}.
What we are uncovering here is that a basic measure of
the situation --- the blunder potential, which as we noted above
corresponds to a measure of the {\em danger} inherent in
the underlying chess position --- is arguably playing
a larger role than the players' skill.
This finding also relates to work of Abelson on quantitative
measures in a different competitive domain, baseball, where
he found that a player's batting average accounts for very
little of the variance in their performance in any single at-bat
\cite{abelson-paradox}.

We should emphasize, however, that despite the strong effect of blunder
potential, skill does play a fundamental role
role in our domain, as the analysis of this section has shown.
And in general it is important to take multiple
types of features into account in any analysis of decision-making,
since only certain features may be under our control
in any given application.  For example, we may be able to control the
quality of the people we recruit to a decision,
even if we can't control the difficulty of the decision itself.

\xhdr{The Skill Gradient for Fixed Positions}
Grouping positions together by common $(n(P),b(P))$ values
gives us a rough sense for how the skill gradient
behaves in positions of varying difficulty.
But this analysis still aggregates together a large number of different
positions, each with their own particular properties, and so it
becomes interesting to ask --- how does the empirical blunder rate
vary with Elo rating {\em when we fix the exact position on the board?}

The fact that we are able to meaningfully ask this question is based
on a fact noted in Section \ref{sec:intro}, that many non-trivial
$\leq$$6$-piece positions recur in the FICS data, exactly,
several thousand times.\footnote{To increase the amount of data we have
on each position, we cluster together positions
that are equivalent by symmetry: we can apply a left-to-right reflection of the board, or we can apply a top-bottom reflection of the board (also reversing the colors of the pieces and the side to move),
or we can
do both.  Each of the four resulting positions is equivalent under
the rules of chess.}
For each such position $P$, we have
enough instances to plot the function $f_P(x)$, the rate of
blunders committed by players of rating $x$ in position $P$.

Let us say that the function $f_P(x)$ is {\em skill-monotone}
if it is decreasing in $x$ --- that is, if players of higher rating
have a lower blunder rate in position $P$.
A natural conjecture would be that every position $P$ is skill-monotone,
but in fact this is not the case.
Among the most frequent positions, we find several that
we term {\em skill-neutral}, with $f_P(x)$ remaining
approximately constant in $x$, as well as several that we term {\em skill-anomalous},
with $f_P(x)$ increasing in $x$.
Figure \ref{fig:skill-gradient-position} shows a subset of
the most frequently occurring positions in the FICS data that contains
examples of each of these three types:
skill-monotone, skill-neutral, and skill-anomalous.\footnote{For
readers interested in looking
at the exact positions in question, each position in
Figure \ref{fig:skill-gradient-position} is described in
Forsyth-Edwards notation (FEN) above the panel in which its plot appears.}

The existence of skill-anomalous positions is surprising, since
there is a no {\em a priori} reason to believe that chess as a domain
should contain common situations in which stronger players make more
errors than weaker players.
Moreover, the behavior of players in these particular positions does not seem
explainable by a strategy in which they are deliberately
making a one-move blunder for the sake of the overall game outcome.
In each of the skill-anomalous examples in
Figure \ref{fig:skill-gradient-position}, the player to move
has a forced win, and the position is reduced enough that the worst possible
game outcome for them is a draw under any sequence of moves,
so there is no long-term value
in blundering away the win on their present move.

\subsection{Time}

Finally, we consider our third category of features, the time
that players have available to make their moves.
Recall that players have to make their own decisions about
how to allocate a fixed budget of time across a given number of moves
or the rest of the game.
The FICS data has information about
the time remaining associated with each move in each game,
so we focus our analysis on FICS in this subsection.
Specifically, as noted in Section \ref{sec:features},
FICS games are generally played under extremely rapid conditions, and
for uniformity in the analysis we focus on the most commonly-occurring
FICS time constraint ---
the large subset of games in which each player
is allocated
3 minutes for the whole game.

\begin{figure}[t]
\centering
\includegraphics[width=0.3\textwidth]{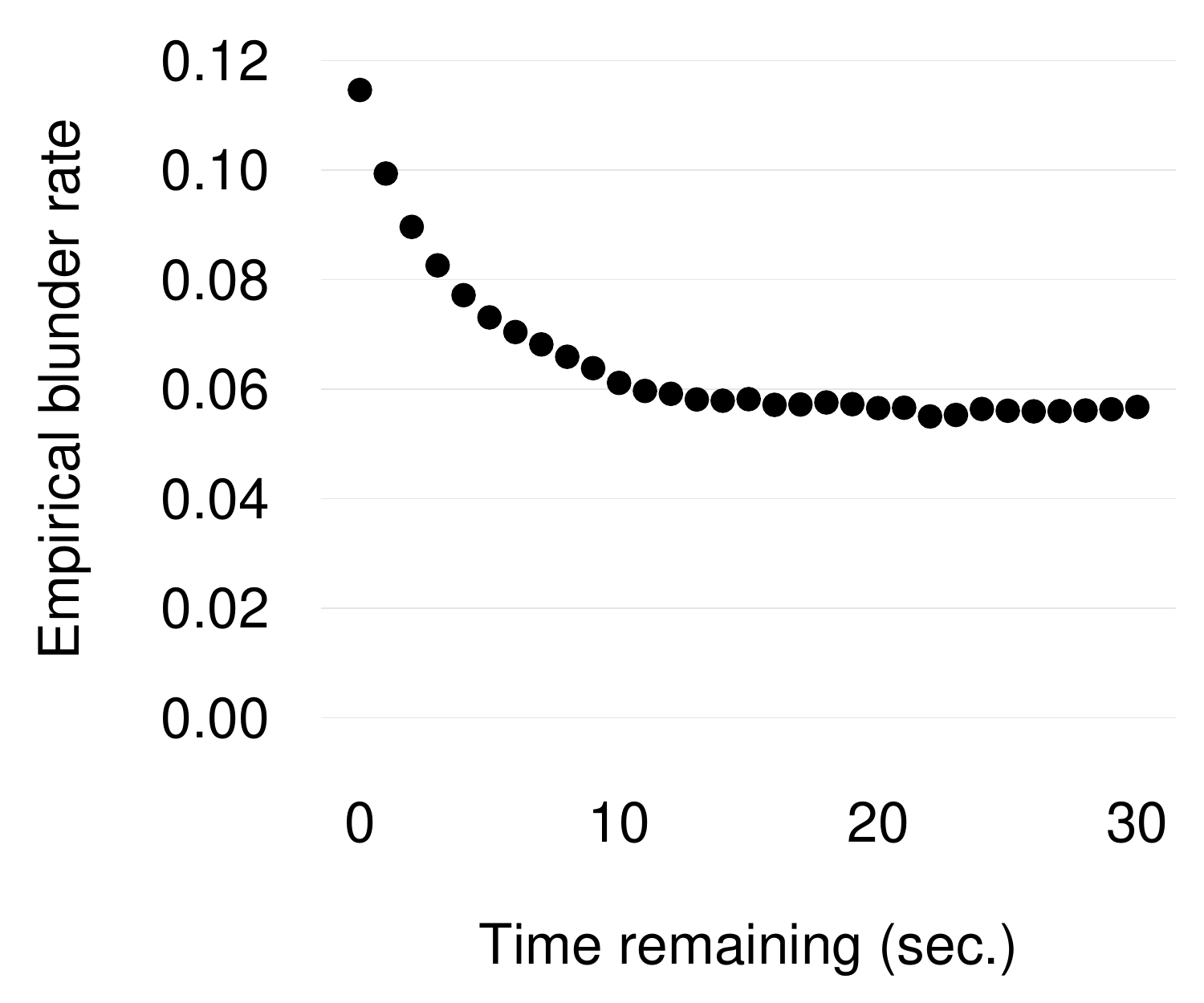} %
\vspace{-2mm}
\caption{The empirical blunder rate as a function of time remaining.}
\label{fig:blunder-rate-time-left}
\end{figure}

As a first object of study, let's define the function $g(t)$ to be
the empirical blunder rate in positions where
the player begins considering their move with $t$ seconds left in the game.
Figure \ref{fig:blunder-rate-time-left} shows a plot of $g(t)$;
it is natural that the blunder rate increases sharply as $t$ approaches
$0$, though it is notable how flat the value of $g(t)$ becomes
once $t$ exceeds roughly 10 seconds.

\begin{figure}[t]
\centering
\vspace{-2mm}
\includegraphics[width=0.47\textwidth]{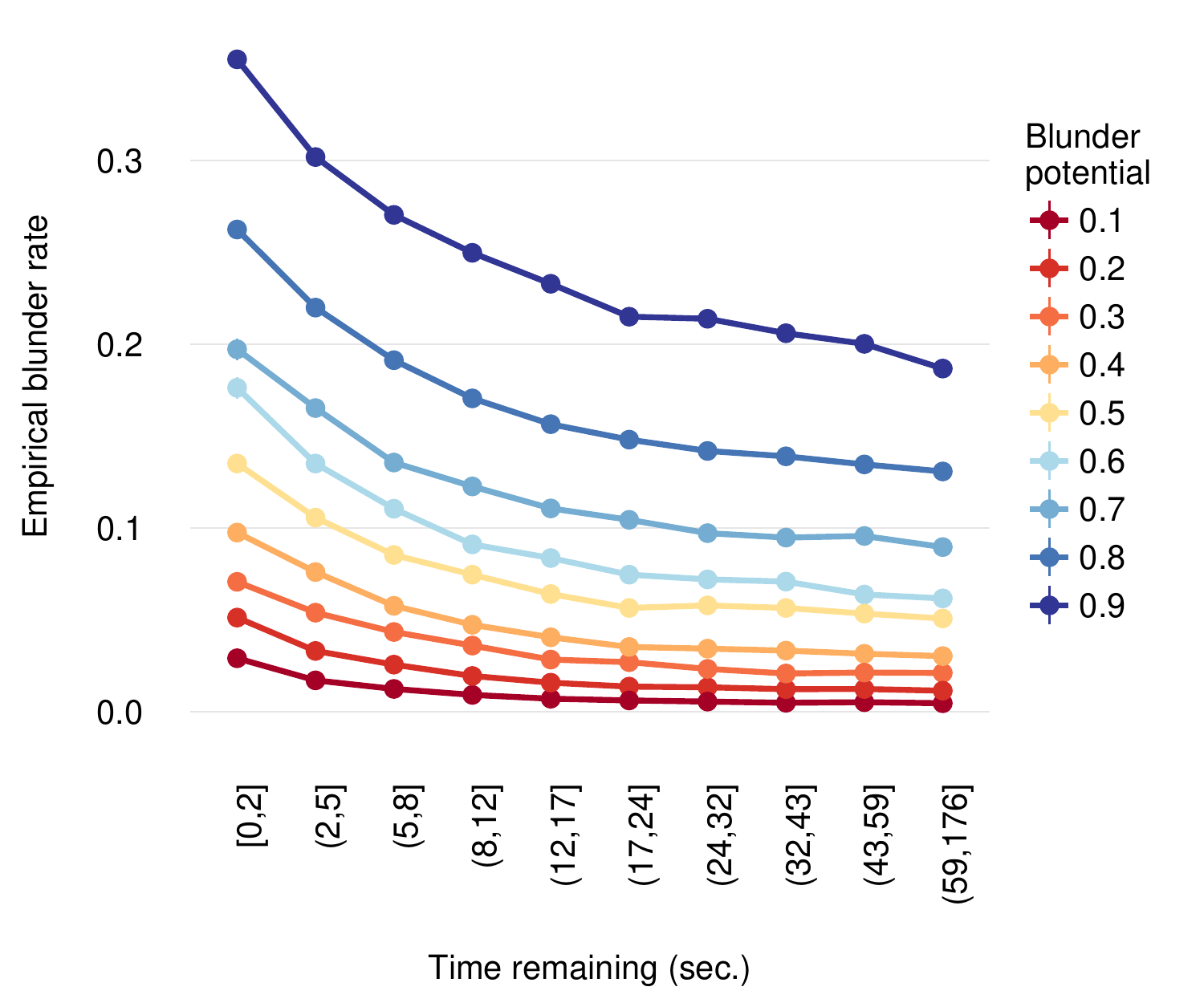}
\vspace{-2mm}
\caption{The empirical blunder rate as a function of the time remaining,
for positions with fixed blunder potential values.}
\label{fig:blunder-rate-potential-time-left}
\vspace{-3mm}
\end{figure}

\begin{figure}[t]
\centering
\hspace*{-0.03\textwidth}
\vspace{-2mm}
\includegraphics[width=0.5\textwidth]{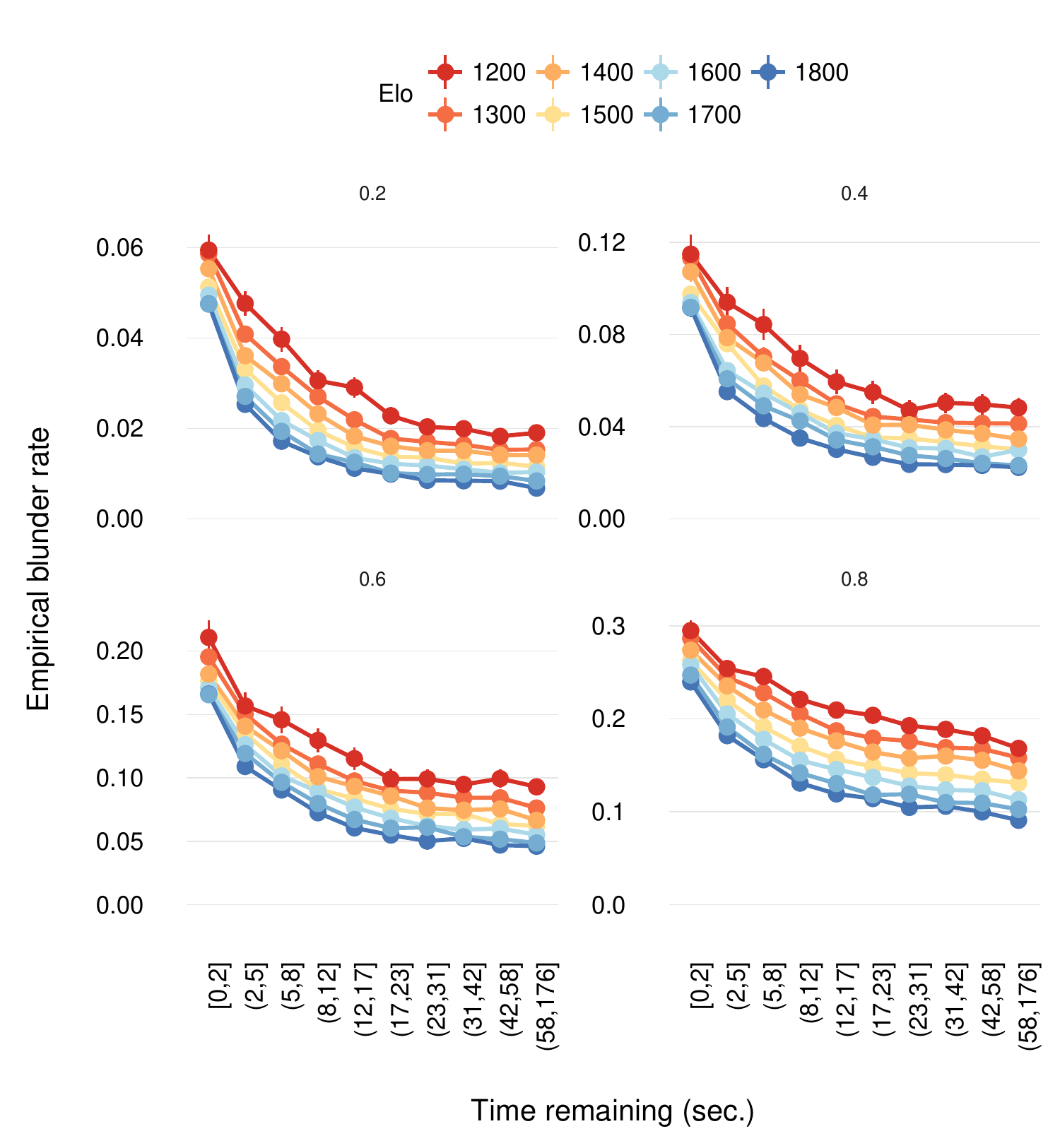}
\vspace{-2mm}
\caption{The empirical blunder rate as a function of time remaining,
fixing the blunder potential and player rating.}
\label{fig:time-gradient}
\vspace{-3mm}
\end{figure}

\xhdr{The Time Gradient}
This plot in Figure \ref{fig:blunder-rate-time-left} can
be viewed as a basic kind of {\em time gradient}, analogous to the
skill gradient, showing the overall improvement in empirical blunder rate
that arises from having extra time available.
Here too we can look at how the time gradient restricted to positions
with fixed blunder potential, or fixed blunder potential and player rating.

We start with Figure \ref{fig:blunder-rate-potential-time-left}, which
shows $g_\beta(t)$, the blunder rate for players within a narrow skill range (1500-1599 Elo) with $t$ seconds remaining in positions with blunder potential $\beta$.
In this sense, it is a close analogue
of Figure \ref{fig:blunder-rate-potential-rating}, which plotted
$f_\beta(x)$, and for values of $t$ above $8$ seconds, it shows
a very similar ``ladder'' structure in which the role
of blunder potential is dominant.
Specifically, for every $\beta$, players are blundering at a lower rate with
$8$ to $12$ seconds remaining at blunder potential $\beta$
than they are with over a minute remaining at blunder potential
$\beta + 0.2$.
A small increase in blunder potential has a more extensive effect
on blunder rate than a large increase in available time.

We can separate the instances further both by blunder potential
and by the rating of the player, via the function $g_{\beta,x}(t)$
which gives the empirical blunder rate with $t$ seconds remaining
when restricted to players of rating $x$ in positions of
blunder potential $\beta$.
Figure \ref{fig:time-gradient} plots these functions, with
a fixed value of $\beta$ in each panel.
We can compare curves for players of different rating,
observing that for higher ratings the curves are steeper:
extra time confers a
greater relative empirical benefit on higher-rated players.
Across panels, we see that
for higher blunder potential the curves become somewhat shallower: more time
provides less relative improvement as the density of possible blunders
proliferates.
But equally or more striking is the fact that all curves retain
a roughly constant shape,
even as the empirical blunder rate climbs
by an order of magnitude from the low ranges of blunder potential
to the highest.

Comparing across points in different panels helps drive home
the role of blunder potential even when considering skill and time
simultaneously.
Consider for example (a) instances in which players rated 1200 (at the low
end of the FICS data) with 5-8 seconds remaining face a position of
blunder potential 0.4, contrasted with (b) instances in which players
rated 1800 (at the high end of the FICS data) with 42-58 seconds remaining
face a position of blunder potential 0.8.
As the figure shows, the empirical blunder rate is lower in instances
of type (a) --- a weak player in extreme time pressure is making blunders
at a lower rate because they're dealing with positions that contain less danger.

\xhdr{Time Spent on a Move}
Thus far we've looked at how the empirical blunder rate
depends on the amount of time remaining in the game.
However, we can also ask how the probability of a blunder varies
with the amount of time the player actually spends considering
their move before playing it.
When a player spends more time on a move, should we predict
they're less likely to blunder (because they gave the move
more consideration) or more likely to blunder (because the
extra time suggests they didn't know what do)?

The data turns out to be strongly consistent with the latter view:
the empirical blunder rate is higher in aggregate for players who
spend more time playing a move.
We find that this property holds across the range of possible values for
the time remaining and the blunder potential, as well as
when we fix the specific position.

\section{Prediction}
\label{sec:prediction}
We've now seen how the empirical blunder rate depends on our three fundamental dimensions: difficulty, the skill of the player, and the time available to them. We now turn to a set of tasks that allow us to further study the predictive power of these dimensions.

\subsection{Greater Tree Depth}
\label{sec:complex}

In order to formulate our prediction methods for blunders, we first extend
the set of features available for studying the difficulty of a position.
Once we have these additional features, we will be prepared to
develop the predictions themselves.

Thus far, when we've considered
a position's difficulty, we've used information
about the player's immediate moves, and then invoked a tablebase
to determine the outcome after these immediate moves.
We now ask whether it is useful for our task to consider longer
sequences of moves beginning at the current position.
Specifically, if we consider all $d$-move sequences beginning at
the current position, we can organize these into a {\em game tree}
of depth $d$ with the current position $P$ as the root, and nodes
representing the states of the game after each possible
sequence of $j \leq d$ moves.
Chess engines use this type of tree as their central structure
in determining which moves to make;
it is less obvious, however, how to make use of these trees
in analyzing blunders by human players, given players' imperfect
selection of moves even at depth 1.

Let us introduce some notation to describe how we use this information.
Suppose our instance consists of position $P$, with
$n$ legal moves, of which $b$ are blunders.
We will denote the moves by $m_1, m_2, ..., m_n$, leading to positions
$P_1, P_2, \ldots, P_n$ respectively, and we'll suppose they are indexed
so that $m_1, m_2, \ldots, m_{n-b}$ are the non-blunders, and
$m_{n-b+1}, \ldots, m_n$ are the blunders.
We write $T_0$ for the indices of the non-blunders
$\{1, 2, \ldots, n-b\}$ and
$T_1$ for the indices of the blunders $\{n-b+1, \ldots, n\}$.
Finally, from each position $P_i$, there are $n_i$ legal
moves, of which $b_i$ are blunders.

The set of all pairs $(n_i,b_i)$ for $i = 1, 2, \ldots, n$ constitutes
a potentially useful source of information in the depth-2 game tree from the
current position.
What might it tell us?

First, suppose that position $P_i$, for $i \in T_1$, is a position
reachable via a blunder $m_i$.
Then if the blunder potential $\beta(P_i) = b_i/n_i$
is large, this means that it may be challenging for the opposing player
to select a move that capitalizes on
the blunder $m_i$ made at the root position $P$;
there is a reasonable chance that the opposing will instead blunder,
restoring the minimax value to something larger.
This, in turn, means that it may be harder for the
player in the root position of our instance to see that move $m_i$,
leading to position $P_i$, is in fact a blunder.
The conclusion from this reasoning is that when the blunder potentials
of positions $P_i$ for $i \in T_1$ are large, it suggests
a larger empirical blunder rate at $P$.

It is less clear what to conclude when there are large blunder potentials
at positions $P_i$ for $i \in T_0$ --- positions reachable by non-blunders.
Again, it suggests that player at the root may have a harder time
correctly evaluating the positions $P_i$ for $i \in T_0$; if they appear
better than they are, it could lead the player to favor these non-blunders.
On the other hand, the fact that these positions are hard to evaluate
could also suggest a general level of difficulty in evaluating $P$,
which could elevate the empirical blunder rate.

There is also a useful aggregation of this information, as follows.
If we define $b(T_1) = \sum_{i \in T_1} b_i$ and
$n(T_1) = \sum_{i \in T_1} n_i$, and analogously for $b(T_0)$ and $n(T_0)$,
then the ratio $\beta_1 = b(T_1)/n(T_1)$ is a kind of aggregate blunder potential
for all positions reachable by blunders, and analogously
for $\beta_0 = b(T_0)/n(T_0)$ with respect to positions reachable by non-blunders.

In the next subsection, we will see that the four quantities
$b(T_1)$, $n(T_1)$, $b(T_0)$, $n(T_0)$ indeed contain useful information
for prediction, particularly when looking at families of
instances that have the same blunder potential at the root position $P$.
We note that one can construct analogous information at greater
depths in the game tree, by similar means,
but we find in the next subsection that
these do not currently provide improvements in prediction performance,
so we do not discuss greater depths further here.

\subsection{Prediction Results}
\label{subsec:prediction}

We develop three nested prediction tasks: in the first task we make
predictions about an unconstrained set of instances; in the second we
fix the blunder potential at the root position; and in the third we
control for the exact position.

\xhdr{Task 1}
In our first task we formulate the basic error-prediction problem:
we have a large collection of human decisions for which we know
the correct answer, and we want to predict whether the decision-maker
will err or not.  In our context, we predict
whether the player to move will blunder, given the position they face
and the various features of it we have derived, how much time they
have to think, and their skill level.
In the process, we seek to understand the relative value of these
features for prediction in our domain.

We restrict our attention to the 6.6 million instances that occurred in
the 320,000 empirically most frequent positions in the FICS dataset.
Since the rate of blundering is low in general, we down-sample the
non-blunders so that half of our remaining instances are blunders and
the other half are non-blunders. This results in a balanced dataset
with 600,000 instances, and we evaluate model performance with accuracy.
For ease of interpretation, we use both logistic regression and
decision trees. Since the relative performance of these two
classifiers is virtually identical, but decision trees perform
slightly better, we only report the results using decision trees here.

Table~\ref{tab:features} defines the features we use for prediction.
In addition the notation defined thus far,
we define: $S=\{\textrm{Elo},\textrm{Opp-elo}\}$ to be the skill
features consisting of the rating of the player and the opponent;
$a(P)=n(P)-b(P)$ for the number of non-blunders in position $P$;
$D_1=\{a(P),b(P),\beta(P)\}$ to be the difficulty features at depth 1;
$D_2=\{a(T_0), b(T_0), a(T_1),\allowbreak b(T_1), \beta_0(P), \beta_1(P)\}$ as the difficulty features at depth 2
defined in the previous subsection; and $t$ as the time remaining.

\begin{table}[t]

\begin{tabular}{p{2cm}p{5.6cm}}
\toprule
Feature & Description \\
\midrule
$\beta(P)$ & Blunder potential ($b(P)/n(P)$)  \\
$a(P)$, $b(P)$ & \# of non-blunders, \# of blunders \\
$a(T_0)$, $b(T_0)$ & \# of non-blunders and blunders available to opponent following a non-blunder \\
$a(T_1)$, $b(T_1)$ & \# of non-blunders and blunders available to \newline opponent following a blunder \\
$\beta_0(P)$ & Opponent's aggregate $\beta$ after a non-blunder \\
$\beta_1(P)$ & Opponent's aggregate $\beta$ after a blunder \\
Elo, Opp-elo & Player ratings (skill level) \\
$t$ & Amount of time player has left in game \\
\bottomrule
\end{tabular}
\caption{Features for blunder prediction.}
\label{tab:features}
\vspace*{-4mm}
\end{table}

In Table~\ref{tab:task1results}, we show the performance of various
combinations of our features. The most striking result is how dominant
the difficulty features are. Using all of them together gives 0.75
accuracy on this balanced dataset, halfway between random guessing and
perfect performance.
In comparison, skill and time are much less informative on this task.
The skill features $S$ only give 55\% accuracy, time left $t$ yields
53\% correct predictions, and neither adds predictive value once
position difficulty features are in the model.

The weakness of the skill and time features is consistent with
our findings in Section~\ref{sec:basic}, but still striking given
the large ranges over which the Elo ratings and time remaining can extend.
In particular, a player rated 1800 will almost always defeat a player
rated 1200, yet knowledge of rating is not providing much predictive
power in determining blunders on any individual move.
Similarly, a player with 10 seconds remaining in the entire game
is at an enormous disadvantage compared to a player with two minutes
remaining, but this too is not providing much leverage for
blunder prediction at the move level.
While these results only apply to our particular domain, it suggests
a genre of question that can be asked by analogy in many domains.
(To take one of many possible examples, one could similarly ask about
the error rate of highly skilled drivers in difficult conditions
versus bad drivers in safe conditions.)

Another important result is that most of the predictive power comes
from depth 1 features of the tree. This tells us the immediate
situation facing the player is by far the most informative feature.

Finally, we note that the prediction results for the GM data (where
we do not have time information available) are closely analogous;
we get a slightly higher accuracy of $0.77$, and again it comes
entirely from our basic set of difficulty features for the position.

\begin{table}[t]
\begin{tabular}{p{4.6cm}P{3cm}}
\toprule
Model & Accuracy \\
\midrule
Random guessing & 0.50 \\
$\beta(P)$ & 0.73  \\
$D_1$ & 0.73 \\
$D_2$ & 0.72 \\
$D_1 \cup D_2$ & \textbf{0.75} \\
$S$ & 0.55 \\
$S \cup D_1 \cup D_2$ & \textbf{0.75} \\
$\{t\}$ & 0.53 \\
$\{t\} \cup D_1 \cup D_2$ & \textbf{0.75} \\
$S \cup \{t\} \cup D_1 \cup D_2$ & \textbf{0.75} \\
\bottomrule
\end{tabular}
\caption{Accuracy results on Task 1.}
\label{tab:task1results}
\vspace*{-1mm}
\end{table}

\xhdr{Human Performance on a Version of Task 1}
Given the accuracy of algorithms for Task 1,
it is natural to consider how this compares to the performance of
human chess players on such a task.

To investigate this question, we developed a version of Task 1 as a web app quiz and promoted it
on two popular Internet chess forums.  Each quiz question provided a
pair of $\leq$6-piece instances with White to move,
each showing the exact position on the board, the ratings of the two
players, and the time remaining for each.  The two instances were chosen from the FICS data with the property that White
blundered in one of them and not the other, and the quiz question was
to determine in which instance White blundered.

In this sense, the quiz is a different type of chess problem from
the typical style, reflecting the focus of our work here:
rather than ``White to play and win,'' it asked ``Did White blunder in
this position?''.
Averaging over approximately 6000 responses to the quiz from
720 participants, we find an accuracy of $0.69$, non-trivially better
than random guessing but also non-trivially below our model's
performance of $0.79$.\footnote{Note that the model performs slightly better here than in our basic formulation
of Task 1, since there instances were not presented in pairs but simply
as single instances drawn from a balanced distribution of positive
and negative cases.}

The relative performance of the prediction algorithm and the human
forum participants forms an interesting contrast, given that the human
participants were able to use domain knowledge about properties of
the exact chess position while the algorithm is achieving
almost its full performance from a single number -- the blunder potential ---
that draws on a tablebase for its computation.
We also investigated the extent to which the guesses made by human
participants could be predicted by an algorithm; our
accuracy on this was in fact lower than for the
blunder-prediction task itself,
with the blunder potential again serving as the most important feature for
predicting human guesses on the task.

\vspace{-1mm}
\xhdr{Task 2}
Given how powerful the depth 1 features are, we now control for $b(P)$
and $n(P)$ and investigate the predictive performance of our features
once blunder potential has been fixed. Our strategy on this task is
very similar to before: we compare different groups of features
on a binary classification task and use accuracy as our measure. These
groups of features are: $D_2$, $S$, $S \cup D_2$, $\{t\}$, $\{t\} \cup D_2$,
and the full set $S\cup \{t\} \cup D_2$. For each of these models, we have
an accuracy score for every $(b(P),n(P))$ pair. The relative
performances of the models are qualitatively similar across all
$(b(P),n(P))$ pairs: again, position difficulty dominates time and
rating, this time at depth 2 instead of depth 1. In all cases, the
performance of the full feature set is best (the mean accuracy is
0.71), but $D_2$ alone achieves 0.70 accuracy on average. This further underscores the
importance of position difficulty.

Additionally, inspecting the decision tree models reveals a very
interesting dependence of the blunder rate on the depth 1 structure of
the game tree. First, recall that the most frequently
occurring positions in our datasets have either $b(P)=1$ or $b(P) =
n(P)-1$. In so-called ``only-move'' situations, where there is only
one move that is not a blunder, the dependence of blunder rate on
$D_2$ is as one would expect: the higher the $b(T_1)$ ratio, the more
likely the player is to blunder. But for positions with only
one blunder, the dependence reverses: blunders are
{\em less} likely with higher $b(T_1)$ ratios.
Understanding this latter
effect is an interesting open question.

\vspace{-1mm}
\xhdr{Task 3}
Our final prediction question is about the degree to which time and
skill are informative once the position has been fully controlled for.
In other words, once we understand
everything we can about a position's difficulty, what can we learn
from the other dimensions? To answer this question, we set up a final
task where we fix the position completely, create a balanced dataset
of blunders and non-blunders, and consider how well time and skill
predict whether a player will blunder in the position or not. We do
this for all 25 instances of positions for which there are over 500
blunders in our data.  On average, knowing the rating of the player
alone results in an accuracy of 0.62, knowing the
times available to the player and his opponent yields 0.54,
and together they give 0.63. Thus once difficulty has been completely
controlled for, there is still substantive predictive power in skill and
time, consistent with the notion that all three dimensions are
important.

\section{Discussion}
\label{sec:discussion}

We have used chess as a model system to investigate the types
of features that help in analyzing and predicting error
in human decision-making.
Chess provides us with a highly instrumented domain in which
the time available to and skill of a decision-maker are often recorded,
and, for positions with few pieces, the set of
optimal decisions can be determined computationally.

Through our analysis we have seen that the inherent difficulty of
the decision, even approximated simply by the proportion of available
blunders in the underlying position, can be a more powerful source
of information than the skill or time available.
We have also identified a number of other phenomena, including
the ways in which players of different skill levels
benefit differently, in aggregate, from easier instances or more time.
And we have found, surprisingly, that there exist {\em skill-anomalous}
positions in which weaker players commit fewer errors than stronger players.

We believe there are natural opportunities to apply the paper's
framework of skill, time, and difficulty to a range of settings
in which human experts make a sequence of decisions, some of which
turn out to be in error.
In doing so, we may be able to differentiate between domains
in which skill, time, or difficulty emerge as the dominant
source of predictive information.
Many questions in this style can be asked.
For a setting such as medicine, is the experience of
the physician or the difficulty of the case a more important
feature for predicting errors in diagnosis?
Or to recall an analogy raised in the previous section,
for micro-level mistakes in a human task such as driving, we think
of inexperienced and distracted drivers as a major source of risk,
but how do these effects compare to the presence of
dangerous road conditions?

Finally, there are a number of interesting further avenues for
exploring our current model domain of chess positions via tablebases.
One is to more fully treat the domain as a competitive activity
between two parties.
For example, is there evidence in the kinds of positions we study
that stronger players are not only avoiding blunders, but also
steering the game toward positions that have higher blunder
potential for their opponent?
More generally, the interaction of competitive effects with
principles of error-prone decision-making can lead to a rich
collection of further questions.

{\small\xhdr{Acknowledgments} We thank Tommy Ashmore for valuable discussions on chess engines and human chess performance, the {\url{ficsgames.org}} team for providing the FICS data, Bob West for help with web development, and Ken Rogoff, Dan Goldstein, and S\'{e}bastien Lahaie for their very helpful feedback. This work has been supported in part by a Simons Investigator Award, an ARO MURI grant, a Google Research Grant, and a Facebook Faculty Research Grant.}

\end{document}